\documentclass[letterpaper, 10 pt, conference]{ieeeconf}  %

\IEEEoverridecommandlockouts                              %

\usepackage{graphicx} %
\usepackage{amsmath} %
\usepackage{amssymb}  %
\usepackage{url}  %
\usepackage{color}  %
\usepackage{hyperref} %
\usepackage{subcaption}
\usepackage{acronym}
\usepackage{siunitx}

\usepackage{atbegshi}

\newcommand{\transform}[3]{
{^{#2}\mathbf{#1}_{#3}}
}
\newcommand{\point}[2]{
{^{#2}\mathbf{#1}}
}
\newcommand{\norm}[1]{
\lVert #1 \rVert
}

\newacro{esdf}[ESDF]{Euclidean Signed Distance Field}
\newacro{com}[COM]{center of mass}
\newacro{fasm}[FASM]{Force-Angle Stability Margin}
\newacro{ode}[ODE]{Open Dynamics Engine}
\newacro{hpp}[HPP]{Heightmap Pose Prediction}
\newacro{nist}[NIST]{National Institute of Standards and Technology}
\newacro{rrl}[RRL]{Rescue Robot League}

\title{\LARGE \bf
Accurate Pose Prediction on Signed Distance Fields for Mobile Ground Robots in Rough Terrain
}

\author{Martin Oehler and Oskar von Stryk%
\thanks{All authors are with the Simulation, Systems Optimization and Robotics Group, Technical University of Darmstadt, Germany.\linebreak
{\tt\small \{oehler, stryk\}@sim.tu-darmstadt.de}}%
\thanks{Research presented in this paper has been supported by Nexplore within the AICO Collaboration Lab and in parts by the German Federal Ministry of Education and Research (BMBF) within the DRZ project (grant no. 13N16475), and the KIARA project (grant no. 13N16274), and by the LOEWE initiative (Hesse, Germany) within the emergenCITY center.
\newline 979-8-3503-8111-5/23/\$31.00 \textcopyright 2023 IEEE
\newline DOI: 10.1109/SSRR59696.2023.10499944}%
}

\AtBeginShipoutFirst{\raisebox{1cm}{\parbox{0.9\textwidth}{\centering\normalsize Preprint of the paper which appeared in:\\
            \large IEEE International Symposium on Safety, Security, and Rescue Robotics (SSRR) 2023}}}

\begin{document}

\maketitle
\thispagestyle{empty}
\pagestyle{empty}

\begin{abstract}
Autonomous locomotion for mobile ground robots in unstructured environments such as waypoint navigation or flipper control requires a sufficiently accurate prediction of the robot-terrain interaction.
Heuristics like occupancy grids or traversability maps are widely used but limit actions available to robots with active flippers as joint positions are not taken into account.
We present a novel iterative geometric method to predict the 3D pose of mobile ground robots with active flippers on uneven ground with high accuracy and online planning capabilities. This is achieved by utilizing the ability of signed distance fields to represent surfaces with sub-voxel accuracy.
The effectiveness of the presented approach is demonstrated on two different tracked robots in simulation and on a real platform. Compared to a tracking system as ground truth, our method predicts the robot position and orientation with an average accuracy of $\qty{3.11}{\centi\meter}$ and $\qty{3.91}{\degree}$, outperforming a recent heightmap-based approach.
The implementation is made available as an open-source ROS package.

\end{abstract}

\section{INTRODUCTION}

Mobile ground robots can support humans in a wide range of applications. In disaster response~\cite{kruijff2021german} and planetary exploration~\cite{otsu2020fast} they enter hazardous areas and provide a remote presence in environments dangerous to humans. In an industrial application such as construction site monitoring~\cite{becker20223d}, mobile robots automate tedious and repetitive tasks. These environments have in common that they are unstructured and require the traversal of challenging uneven terrain.

Tracked robots are well suited to these environments because of their ability to negotiate rough terrain. Many platforms can additionally reconfigure their kinematics to further improve their capability of overcoming obstacles, e.g., by changing the shape of the tracks with active flippers or by shifting the \acl{com} with a heavy manipulator arm.
However, the additional degrees of freedom make teleoperation of the robot more challenging, therefore, increasing the mental load of the operator and the risk of errors.

Autonomous locomotion capabilities such as path planning~\cite{norouzi2017planning} or whole-body planning~\cite{oehler2020optimization} can alleviate the operator of stress but as part of the planning process, a prediction about the robot-terrain interaction is required.
A common approach is the approximation of this interaction via heuristics such as an occupancy grid~\cite{elfes1989using} or traversability map~\cite{wermelinger2016navigation}. While this works well in semi-structured terrain, the approximation has to be chosen conservatively to prevent accidents and therefore limits the actions available to the robot significantly. Moreover, heuristics are not optimal for articulated robots because they do not take the current joint positions into account.
Pose prediction approaches, also known as robot settling, can provide a more accurate model of the robot-terrain interaction by estimating the 3D robot position and orientation on the ground.

\begin{figure}[t]
    \centering
    \includegraphics[width=\linewidth]{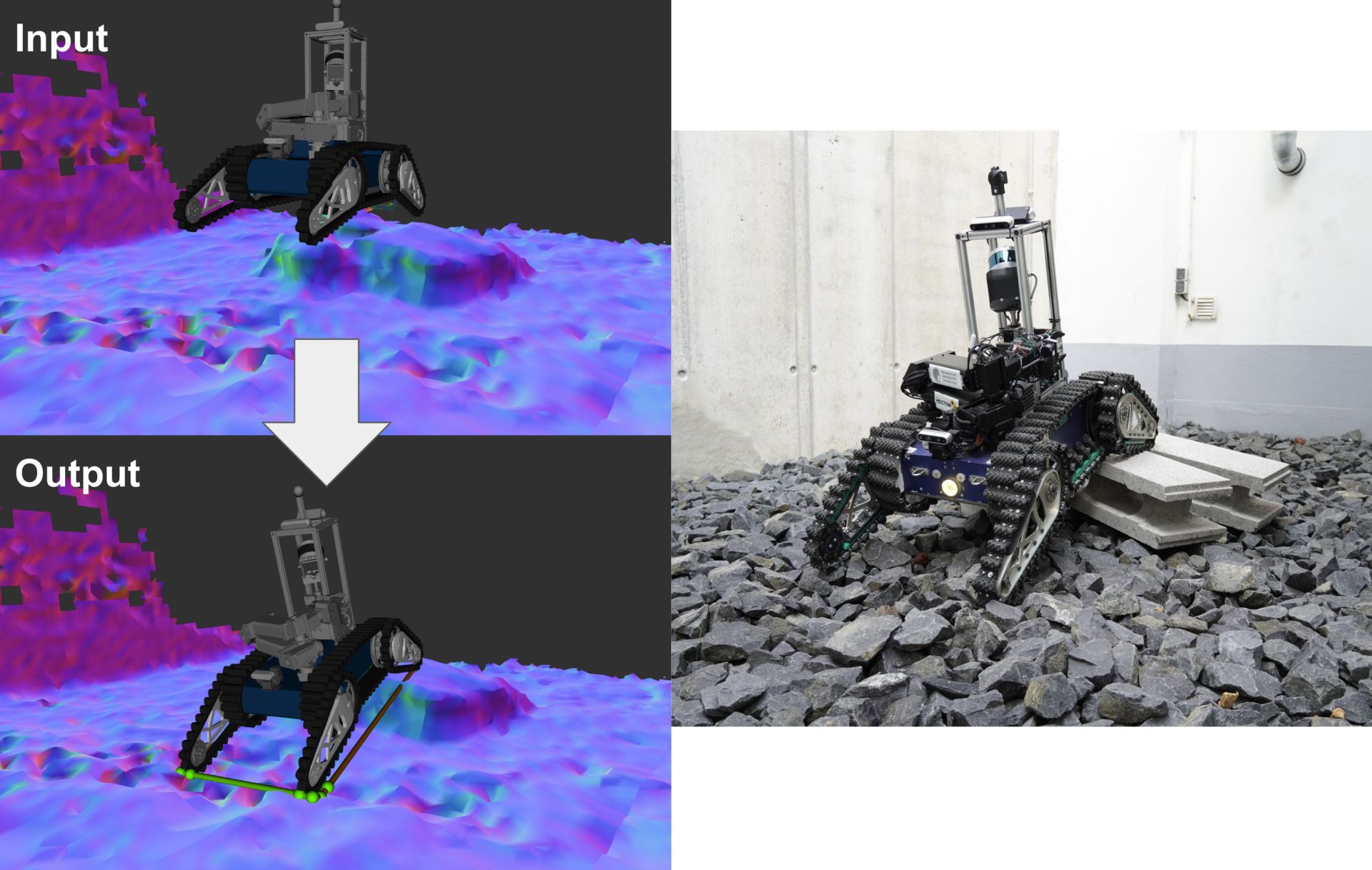}
    \caption{Based on an SE(2) pose, the joint configuration and an \ac{esdf} of the environment (top-left), the 3D pose and terrain interaction are predicted (bottom-left). The photo on the right shows the robot Asterix on the same terrain for comparison.}
    \label{fig:approach_front_page}
\end{figure}

We propose a novel pose prediction method for mobile ground robots based on \acfp{esdf}~\cite{oleynikova2017voxblox} that takes the joint configuration into account (see Fig~\ref{fig:approach_front_page}).
Moreover, the 3D representation allows for the support of multi-level environments. The implementation is available as open-source for the robot framework ROS\footnote{\url{https://github.com/tu-darmstadt-ros-pkg/sdf_contact_estimation}}.

We demonstrate that our approach generalizes to different ground robots by evaluating on two platforms in simulation. Additionally, we perform experiments on the real robot using a tracking system for ground truth localization data and show our approach to be fast enough for online planning. The evaluation scenario is based on the RoboCup \ac{rrl} competition\footnote{\url{https://rrl.robocup.org/}} arenas that are designed to simulate unstructured environments found in real rescue missions. The dataset and code are publicly available to facilitate the reproduction of our results.

\section{RELATED WORK}

Various methods have been developed to estimate the robot-terrain interaction for planning algorithms. An over\-view and classification of prior work on traversability estimation methods is given in~\cite{papadakis2013terrain}.
A common approach to approximate the robot-terrain interaction is using heuristics.
In \cite{elfes1989using}, an occupancy grid is proposed for robot navigation. The 2D grid representation stores the probability of being occupied in each cell and therefore simplifies the problem to a binary decision whether a surface is traversable or not.
This limitation is addressed by the authors of \cite{wermelinger2016navigation}. They propose the traversability map which introduces a continuous traversability value based on the local ground geometry, encoding factors such as roughness, step height, inclined slopes, or gaps.
While being fast to compute, heuristics only give a conservative estimate of the robot-terrain interaction. They can not represent situations in which a robot may traverse the terrain in one orientation but not in another and do not take the joint configuration into account. Pose prediction approaches address these shortcomings by virtually settling the robot on the ground using a map of the environment. Following \cite{brunner2015design}, we categorize prior work into physics simulation, geometric methods, and learning approaches.

Physics simulations compute forces and accelerations over time to drop the robot under the influence of gravity onto the surface. In \cite{norouzi2017planning}, the physics engine \ac{ode} has been used to simulate poses for a combined path and joint configuration planning. While no timings have been given by the authors, a general physics engine is slow as also shown by \cite{brunner2015design}. The authors of \cite{jun2016pose} propose a specialized approach that formulates the contact problem as a linear complementary problem (LCP) which is solved using Lemke's method. A similar approach is taken by \cite{ma2019pose}. They interpolate the terrain with a cubic B-patch and simulate the robot until gravitational force and contact forces reach a static equilibrium. The contact forces are computed using a singular value decomposition (SVD) with a claimed 10-times speed up over the previous LCP approach.
Geometric methods estimate the robot-terrain interaction directly without simulating time steps. Hence, they are typically faster. In \cite{thakker2021autonomous}, the robot is settled on ground points at a given query pose by fitting a plane on the cloud below the robot footprint. Several traversability metrics are computed from the settled pose and surface cloud. The authors of \cite{otsu2020fast} propose a fast approximate settling for a Mars rover with articulated suspension. They infer the worst-case vehicle configuration from upper and lower bounds for each wheel height and assign metrics such as ground clearance. A more general approach is taken by \cite{brunner2015design}. They propose a fast iterative method that finds the robot pose and contact points based on a geometric model of the robot and a least-squares approximation of the ground. The quality of the estimation depends on how well the terrain below the robot is approximated by a plane.
The authors of \cite{fabian2020pose} avoid this issue by settling the robot directly on a heightmap. By introducing the concept of robot heightmaps, they reduce the pose prediction problem to a sequence of image operations for a fast and accurate computation of poses and contacts with a similar speed to \cite{brunner2015design}. However, using heightmaps as the model for robot and environment comes with the drawbacks of a coarse representation of vertical edges and robot features due to the cell resolution and the limitation to planning problems on a single level in the environment.
\cite{vsalansky2021pose} proposes a learning-based approach that uses self-supervised learning to predict the shape of the supporting terrain. They define the pose estimation problem as the minimization of potential energy and train a pose regressor network that predicts the robot pose given the heightmap.

In the following section, we present a novel iterative geometric approach to predict the robot-terrain interaction. The method achieves fast computation times with consistent results because it is not limited to time steps as would be the case for physics simulation approaches.
By directly modeling the kinematics of the robot in a robot-agnostic manner, the pose prediction takes the current joint configuration into account and generalizes to other robot platforms.
We achieve a high prediction accuracy by modeling the environment with the voxel-based \ac{esdf}. In this representation, each cell encodes the Euclidean distance to the closest surface, therefore enabling the localization of surfaces with sub-voxel accuracy. Moreover, the 3D representation allows for the application in multi-level environments.

\section{METHOD}

\begin{figure*}[t]
    \centering
    \includegraphics[width=\linewidth]{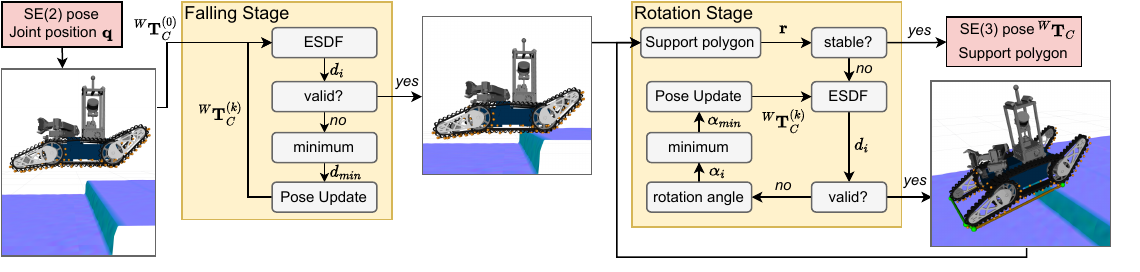}
    \caption{Overview of the pose prediction algorithm. The input pose is used by the Falling Stage to find the first contact with the ground. The Rotation Stage repeatedly rotates the robot about the least stable axis until a stable state is found. Contact candidates are visualized as orange points and the support polygon as green points.}
    \label{fig:algorithm_overview}
\end{figure*}

We present an iterative geometric method to predict the terrain interaction of mobile ground robots. For a given SE(2) position and heading $\left(x, y, \phi\right)$ and joint configuration $\mathbf{q}$, the z position as well as roll and pitch orientation of the SE(3) pose are predicted that are the result of the interaction between robot and terrain. Because this method supports 3D environments with multiple levels, additionally, a coarse estimate for the z position is needed.
The environment is represented by an \ac{esdf} $\Phi\left(\mathbf{p}\right) = d$ that returns the Euclidean distance $d$ to the closest surface of a point in space $\mathbf{p}$. The distances of points in free space have a positive sign while points inside objects are given a negative sign.
The tracks and chassis of the robot are sampled with uniformly distributed contact point candidates.
We assume that the robot and the environment are rigid and that no slippage between robot and surface occurs.

Figure~\ref{fig:algorithm_overview} gives an overview of the algorithm. In the following, the individual steps are briefly described.  The robot pose $\transform{T}{W}{C}$ is given by the transformation between the gravity-aligned world frame $W$ and \acs{com} frame $C$ of the robot. The static, stable robot pose is predicted in an iterative process based on the SE(2) pose which is used as the initial guess for the SE(3) pose $\transform{T}{W}{C}^{(0)}$. First, the robot is dropped from above the ground until the first contact in the Falling Stage. Afterwards, in the Rotation Stage, the robot is repeatedly rotated around the least stable axis of the support polygon until it is in a stable state.

\subsection{Falling Stage}
In this stage, the first contact with the ground is found by translating the robot along the z-axis of the world frame. The translation distance is determined by the minimum distance of any part of the robot to the ground $d_{min}^*$. While the distance to the closest surface $d_{min}$ can be retrieved from the \ac{esdf}, it is unknown whether $d_{min}$ is the distance to the ground or a closer wall. However, it can not be greater than the distance to the ground: $d_{min} \leq d_{min}^*$. Therefore, we can translate by $d_{min}$ to get closer to the ground.
This step is repeated until we find a solution in contact but not in collision with the ground.

First, using the kinematic model of the robot and its joint configuration $\mathbf{q}$, the $N$ contact point candidates are computed in the \acs{com} frame. Using our current estimate of the robot pose $\transform{T}{W}{C}^{(k)}$, each candidate point $\mathbf{p}_i$ is transformed to the world frame and the distance $d_i$ to the closest surface is determined using the \ac{esdf}:

\begin{equation} \label{eq:falling_stage_esdf}
    d_i = \Phi\left(\transform{T}{W}{C}^{(k)} \cdot \point{p}{C}_i\right)
\end{equation}

Our current estimated pose is valid if there is at least one contact point with the ground (\ref{eq:falling_stage_contact}) and no contact candidate is below the ground (\ref{eq:falling_stage_valid}):

\begin{align}
    \exists d \in {d_1, ..., d_N}: d < \epsilon \label{eq:falling_stage_contact}\\
    \forall d \in {d_1, ..., d_N}: d \geq 0 \label{eq:falling_stage_valid}
\end{align}

where $\epsilon$ is the distance threshold for contacts. If these conditions are fulfilled, we continue with the Rotation Stage. Otherwise, we perform an iteration by computing the minimum distance as

\begin{equation}
    d_{min} = \min\left(d_1, ..., d_N\right)
\end{equation}

and translating the robot pose by this distance along the $z$-axis of the world frame with the adaptive step size $\Delta h$.

\begin{align}
    \transform{T}{W}{C}^{(k+1)} =& \transform{T}{W}{C}^{(k)} \cdot \mathbf{Trans} (z; -\Delta h^{(k)} \cdot d_{min}) \\
    \Delta h^{(k+1)} =& \Delta h^{(k)} \cdot h_{f}
\end{align}

Note that because points inside objects return negative distances, $d_{min}$ can also be negative and, therefore, move the robot upwards. For this reason, the initial pose estimate can also be inside the ground.
The step size is initialized with $\Delta h^{(0)} = 1$ and reduced each iteration by $h_{f} \in (0, 1)$ to prevent oscillation and ensure convergence. The updated pose estimate is used to perform the next iteration starting with (\ref{eq:falling_stage_esdf}).

\subsection{Rotation Stage}
Based on the first contact found in the Falling Stage, the robot is now rotated around the least stable axis until a stable state is found.
At the start of this phase, there are three possible contact cases depending on the number of contacts that determine how the rotation axis $\mathbf{r}$ is computed.

In case of only one contact point, the rotation axis is orthogonal to the vector between contact point $\mathbf{c}$ and \ac{com} $\mathbf{p}_{com}^{(k)}$, given as the translational part of $\transform{T}{W}{C}^{(k)}$.
It is computed as the cross product between the vector of \acs{com} to contact point and the (normalized) gravity vector:

\begin{align}
    \mathbf{r} = \left(\mathbf{p}_{com}^{(k)} - \mathbf{c}\right) \times (0, 0, -1)^T
\end{align}

In case of two contact points, the rotation axis is simply the line connecting both points:

\begin{equation}
    \mathbf{r} = \mathbf{c}_2 - \mathbf{c}_1
\end{equation}

For the case of three or more contact points, we obtain the convex hull of all contact points, also called support polygon, using Andrew's monotone chain algorithm~\cite{andrew1979another}.
We use the \ac{fasm}~\cite{papadopoulos2000force} to assess the stability margin of each axis of the support polygon $\mathbf{a}_i = \mathbf{c}_i - \mathbf{c}_{i+1}$ and the minimum:

\begin{align}
    \beta_i =& \text{FASM}\left(\mathbf{a}_i\right) \\
    \beta_{min} =& \min\left(\beta_1,...\beta_{N_c}\right)
\end{align}

If the minimum stability is positive $\beta_{min} > 0$, the current robot pose estimate is stable and the algorithm terminates. Otherwise, the axis with the lowest stability margin is set as the rotation axis.

In the next step, the robot is repeatedly rotated around this axis towards the \ac{com} until the next contact with the ground is found.  The rotation angle is determined as the minimum angle until any part of the robot would make contact with the ground. This step is repeated until a valid solution is found, i.e. in contact with the ground but not in collision.

\begin{figure}
    \centering
    \includegraphics{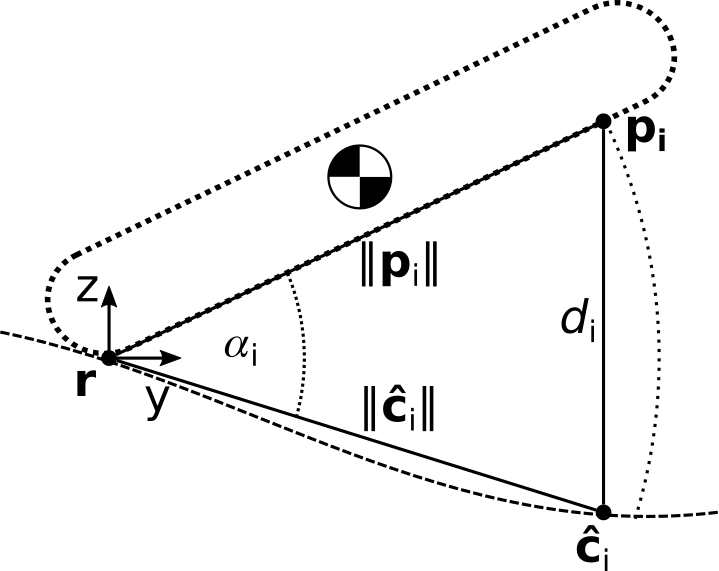}
    \caption{Side-view of the Rotation Stage. The rotation axis $\mathbf{r}$, the contact candidate $\mathbf{p}_i$ and the predicted contact point $\mathbf{\hat{c}}_i$ form a triangle. $d_i$ is the distance to the closest surface, retrieved from the \ac{esdf}. Rotating the robot by $\alpha_i$ would result in a contact.}
    \label{fig:rotation_stage_triangle}
\end{figure}

As in the beginning of the Falling Stage, we start by retrieving the distances of our contact candidates to the closest surface using our latest pose estimate~(\ref{eq:falling_stage_esdf}).
We define a new frame $R$ such that its x-axis is aligned with the rotation axis $\mathbf{r}$, the z-axis is gravity-aligned and the y-axis forms a right-handed coordinate system pointing to the side of the \ac{com}.
Each contact candidate is transformed to this frame. Any candidates that are part of the rotation axis or fall on the negative side of the y-axis are excluded from the following steps. Before an iteration is performed, we check if the current pose estimate fulfills the previously introduced validity constraints~(\ref{eq:falling_stage_contact}), (\ref{eq:falling_stage_valid}). If the solution is valid, we continue with the next Rotation Stage by determining a new rotation axis. Otherwise, we perform the next iteration of this Rotation Stage. Using the distance to the closest surface $d_i$, we predict a possible contact point $\mathbf{\hat{c}}_i$ on the ground (see Fig.~\ref{fig:rotation_stage_triangle}). Note that this introduces a small error as indicated by the dotted circle. In a side-view, the rotation axis $\mathbf{r}$, contact candidate $\mathbf{p}_i$ and predicted contact $\mathbf{\hat{c}}_i$ form a triangle.

For each contact point candidate, we can compute the angle $\alpha_i$ that would lead to a contact with the ground using the law of cosine:

\begin{equation}
    \alpha_i = \sigma_i \arccos \left(\frac{\norm{\mathbf{p}_i}^2 + \norm{\mathbf{\hat{c}}_i}^2 - d_i^2}{2 \norm{\mathbf{p}_i} \norm{\mathbf{\hat{c}}_i}} \right)
\end{equation}

The sign of the angle $\sigma_i$ is negative, if the contact candidate is below the ground:

\begin{equation}
    \sigma_i =
\begin{cases}
+ 1, \; d_i \geq 0\\
- 1, \; d_i < 0\\
\end{cases}
\end{equation}

We obtain the minimum angle with:

\begin{equation}
    \alpha_{min} = \min\left(\alpha_1, ..., \alpha_N\right)
\end{equation}

and rotate the robot pose estimate by this angle:

\begin{align}
  \transform{T}{R}{C}^{(k)} =& \left(\transform{T}{W}{R}\right)^{-1} \transform{T}{W}{C}^{(k)} \\
  \transform{T}{W}{C}^{(k+1)} =&  \transform{T}{W}{R} \cdot \mathbf{Rot}(x; \Delta h^{(k)} \cdot \alpha_{min}) \cdot \transform{T}{R}{C}^{(k)} \\
  \Delta h^{(k+1)} =& \Delta h^{(k)} \cdot h_{f}
\end{align}

As in the Falling Stage, we use a decreasing step size $\Delta h$ to prevent oscillations. The updated pose estimate is used to perform the next iteration using the same rotation axis until the validity constraints are fulfilled. The algorithm terminates, when a stable support polygon has been found.

\section{EVALUATION}
\begin{figure}
    \begin{subfigure}[t]{0.48\linewidth}
        \includegraphics[width=\linewidth]{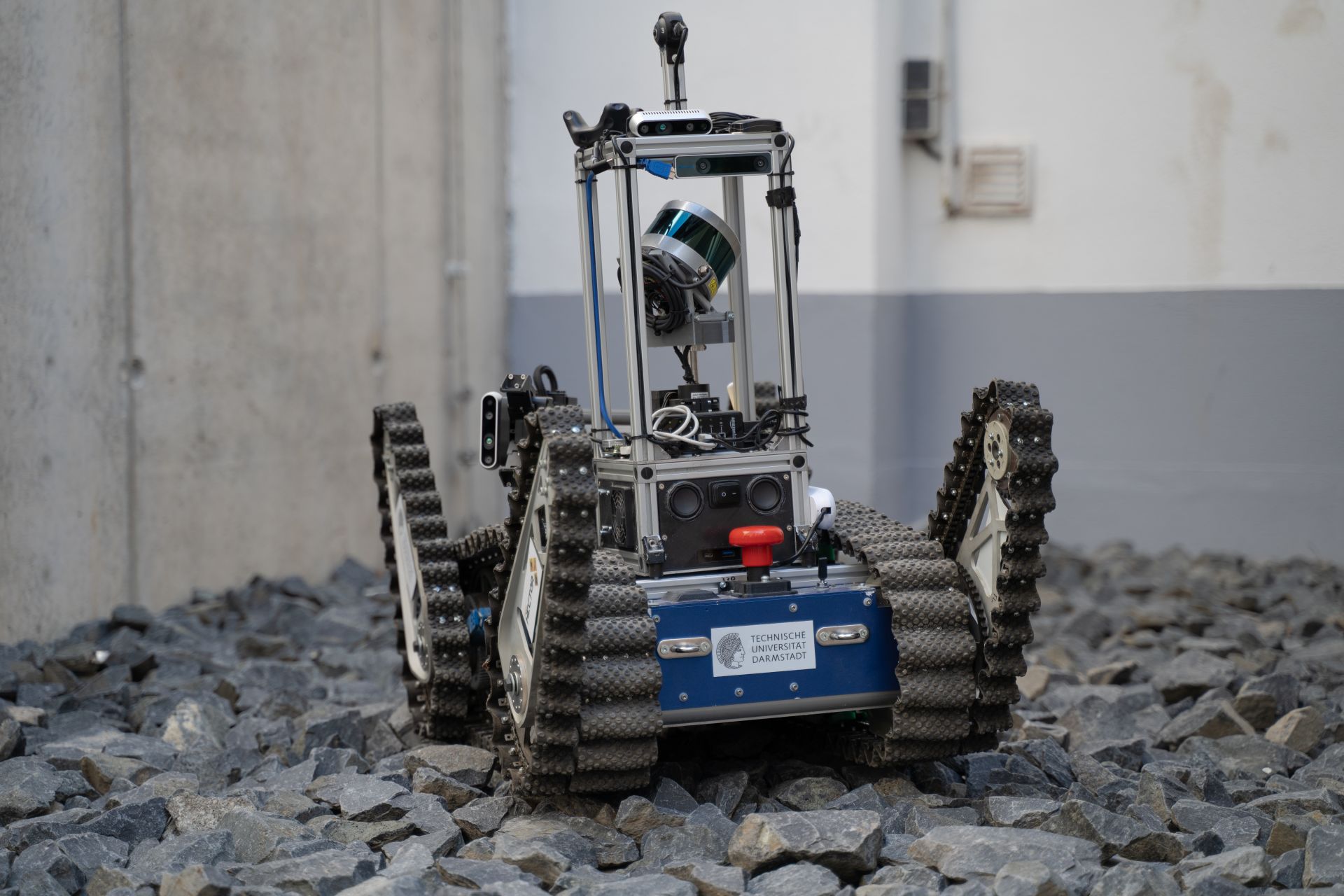}
        \caption{Asterix}\label{fig:asterix}
    \end{subfigure}%
    \begin{subfigure}[t]{0.48\linewidth}
        \includegraphics[width=\linewidth]{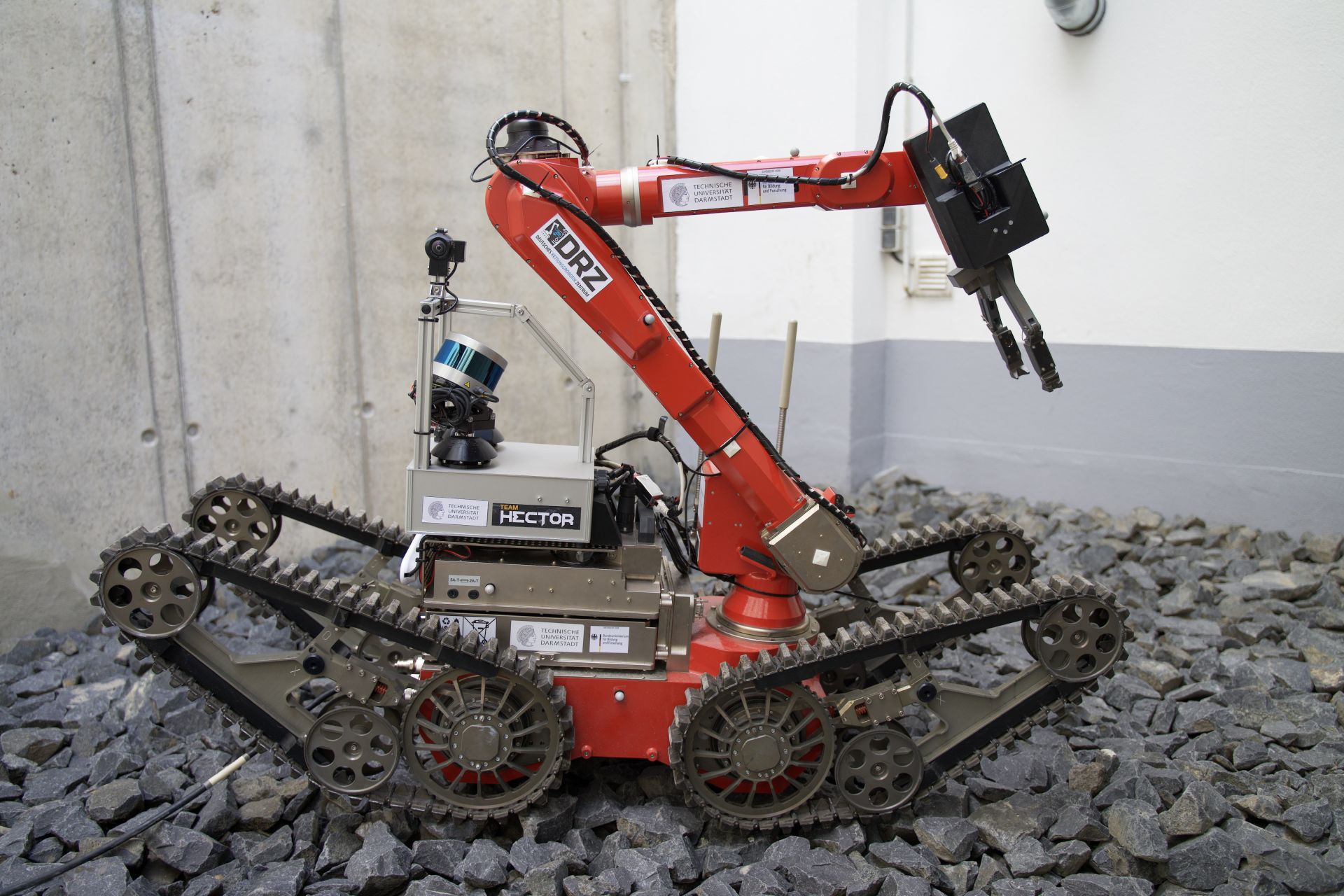}
        \caption{DRZ Telemax}\label{fig:telemax}
    \end{subfigure}%
    \caption{Tracked robot platforms used for evaluation.}
    \label{fig:robots}
\end{figure}

We evaluate the pose prediction accuracy of our approach in multiple scenarios in simulation and on the real robot and compare it to the \ac{hpp} proposed in \cite{fabian2020pose}.
Moreover, by evaluating on two different tracked platforms (see Fig. ~\ref{fig:robots}), we demonstrate the generality of the method. The first robot Asterix is a highly mobile platform with main tracks and coupled flippers on the front and back and a chassis footprint of $72\times\qty{52}{\centi\meter}$. In contrast, the DRZ Telemax robot features a completely different track design with no main tracks, four independent flippers in a unique triangle shape, and a chassis footprint of $81\times\qty{40}{\centi\meter}$.
We use four different arenas of the RoboCup \ac{rrl} for evaluation (see Fig.~\ref{fig:evaluation_scenarios}). Developed by the \ac{nist} and the \ac{rrl}'s technical committee as standard test methods for maneuvering and mobility, they are designed to include unstructured terrain and obstacles also found in real rescue missions.
Each arena offers a different terrain to represent a wide variety of challenges.
The two maneuvering arenas feature easier terrain with a series of double ramps with a height of $\qty{17}{\centi\meter}$ and inclination of $\qty{16}{\degree}$ in the "Continuous Ramps" and three $10\times\qty{10}{\centi\meter}$ bars on flat ground in the "Curb" scenario.
In comparison, the mobility arenas feature more difficult terrain. In the "Hurdles" scenario, the robots have to overcome $\qty{15}{\centi\meter}$ and $\qty{26.5}{\centi\meter}$ steps. The "Elevated Ramps" scenario is composed of boxes with varying heights up to $\qty{35}{\centi\meter}$ and $\qty{16}{\degree}$-slanted tops, therefore, being the most challenging terrain to traverse.

In the following, the process of capturing the evaluation data is explained.
We generate ground truth pose data by driving the robot manually through the terrain and recording its position and orientation in world frame every $\qty{5}{\centi\meter}$. While driving, the flippers are continuously adjusted to overcome the terrain and joint positions are recorded alongside each pose. Onboard distance sensors like Lidar and depth cameras are used to generate a complete map of the environment. The map representations, \ac{esdf} and heightmap, are created with Voxblox~\cite{oleynikova2017voxblox} and elevation\_mapping~\cite{Fankhauser2018ProbabilisticTerrainMapping} respectively, each with a voxel or cell size of $\qty{5}{\centi\meter}$.

We measure the prediction accuracy by comparing the ground truth pose with the predicted pose at each recorded pose along the path.
The ground truth SE(3) pose is reduced to an SE(2) pose by setting roll, pitch and z to $0$ and used as input to the pose prediction.
The position error is given as the Euclidean distance between ground truth and predicted position. The orientation error is the minimum rotation angle that is required to align ground truth and predicted orientation. The error metrics are averaged over all pose samples in each scenario.

To enable the reproduction of our experiments and results, we make our evaluation software\footnote{\url{https://github.com/tu-darmstadt-ros-pkg/hector_pose_prediction_benchmark}} and dataset\footnote{\url{https://tudatalib.ulb.tu-darmstadt.de/handle/tudatalib/3978}} publicly available. The supplementary video\footnote{\url{https://youtu.be/3kHDxPnEtHM}} provides additional visualizations of our results.

\subsection{Simulation}

\begin{figure}[tbp]
    \begin{subfigure}[t]{0.48\linewidth}
        \includegraphics[width=\linewidth,trim={5cm 0 4cm 0},clip]{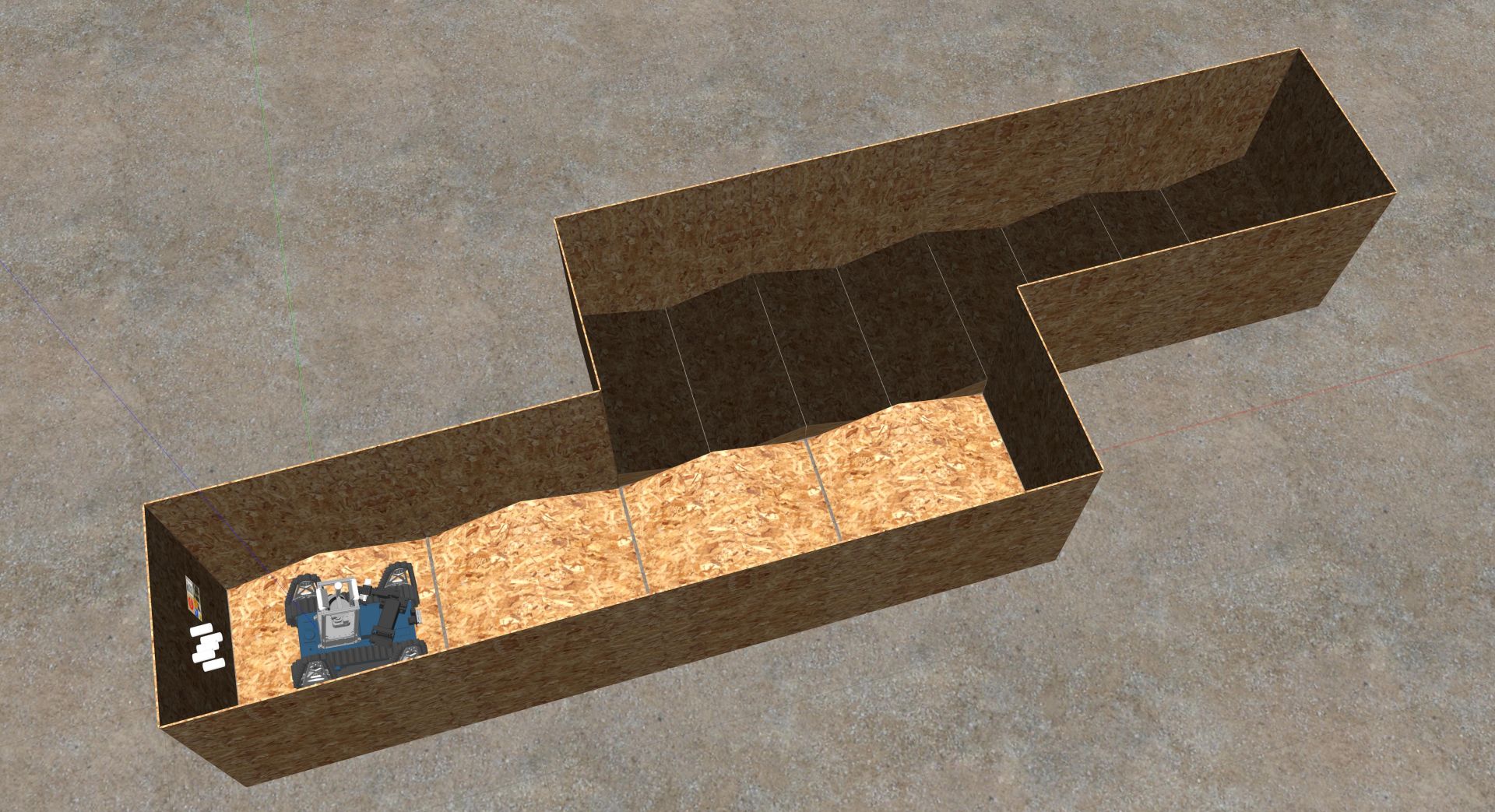}
        \caption{Continuous Ramps}\label{fig:gazebo_continuous_ramps}
    \end{subfigure}%
    \begin{subfigure}[t]{0.48\linewidth}
        \includegraphics[width=\linewidth,trim={5cm 0 4cm 0},clip]{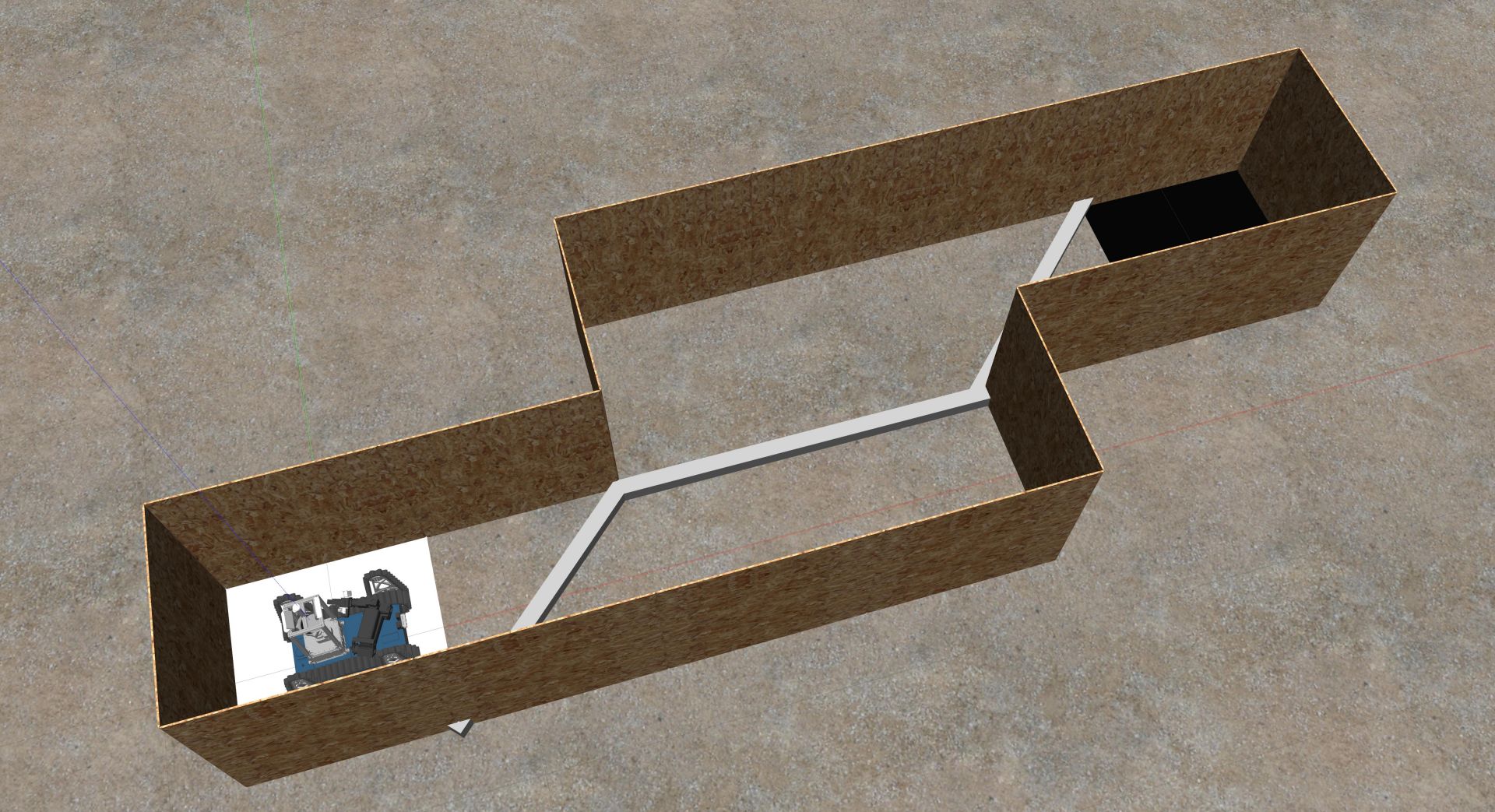}
        \caption{Curb}\label{fig:gazebo_curb}
    \end{subfigure}%
    \\
    \begin{subfigure}[t]{0.48\linewidth}
        \includegraphics[width=\linewidth,trim={5cm 0 4cm 0},clip]{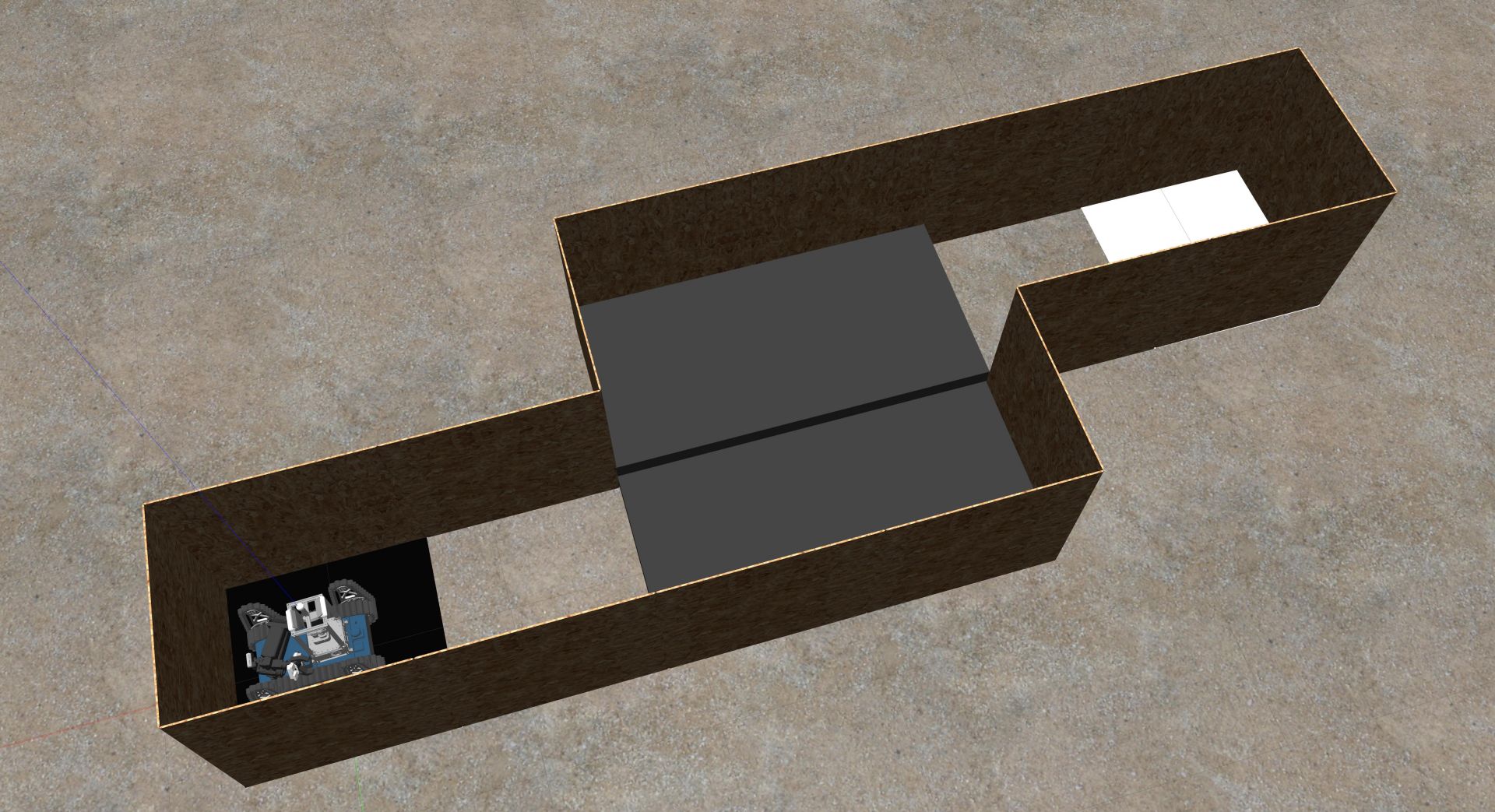}
        \caption{Hurdles}\label{fig:gazebo_hurdles}
    \end{subfigure}%
    \begin{subfigure}[t]{0.48\linewidth}
        \includegraphics[width=\linewidth,trim={5cm 0 4cm 0},clip]{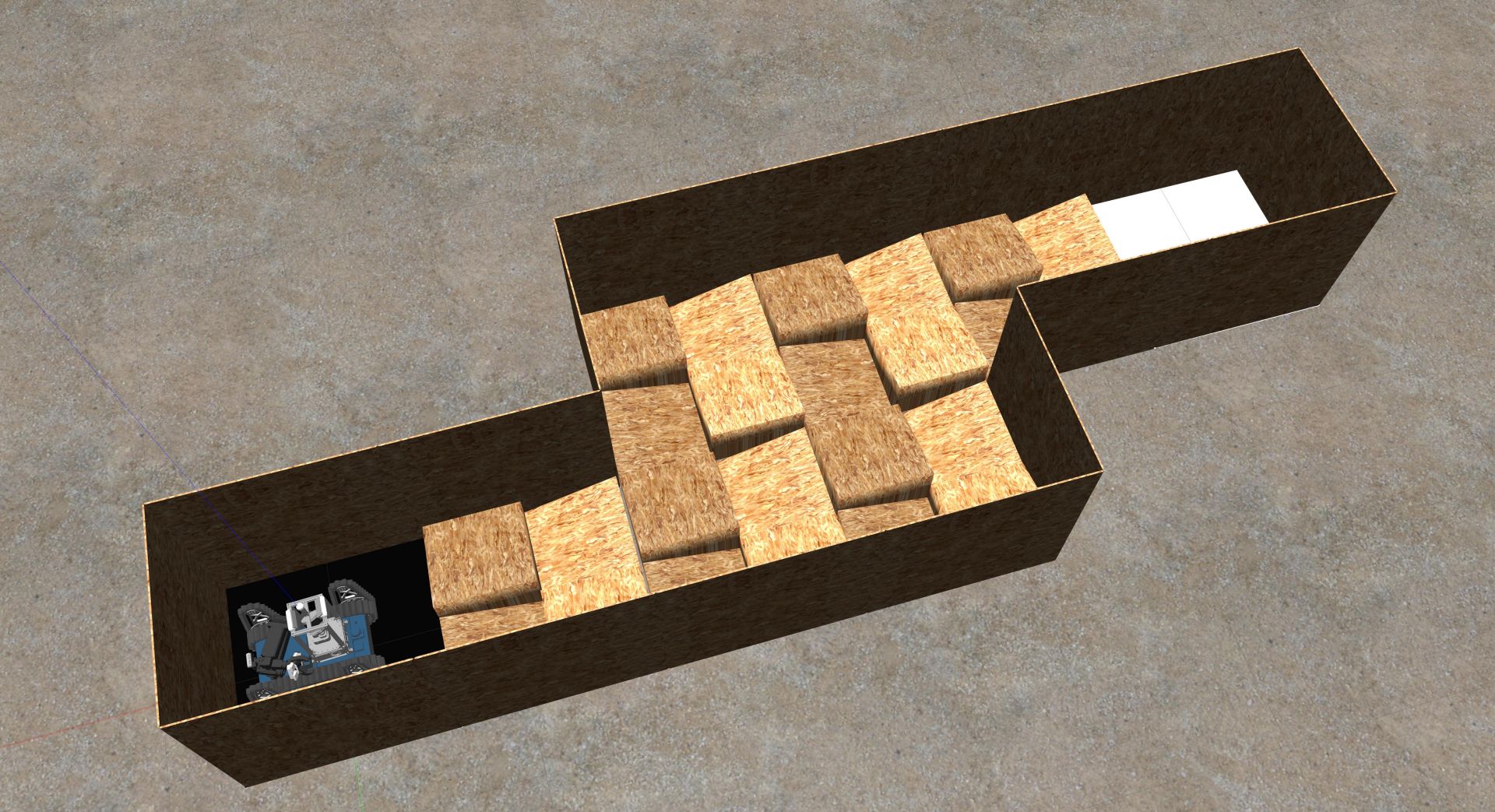}
        \caption{Elevated Ramps}\label{fig:gazebo_elevated_ramps}
    \end{subfigure}%
    \\
    \begin{subfigure}[t]{0.48\linewidth}
        \includegraphics[width=\linewidth]{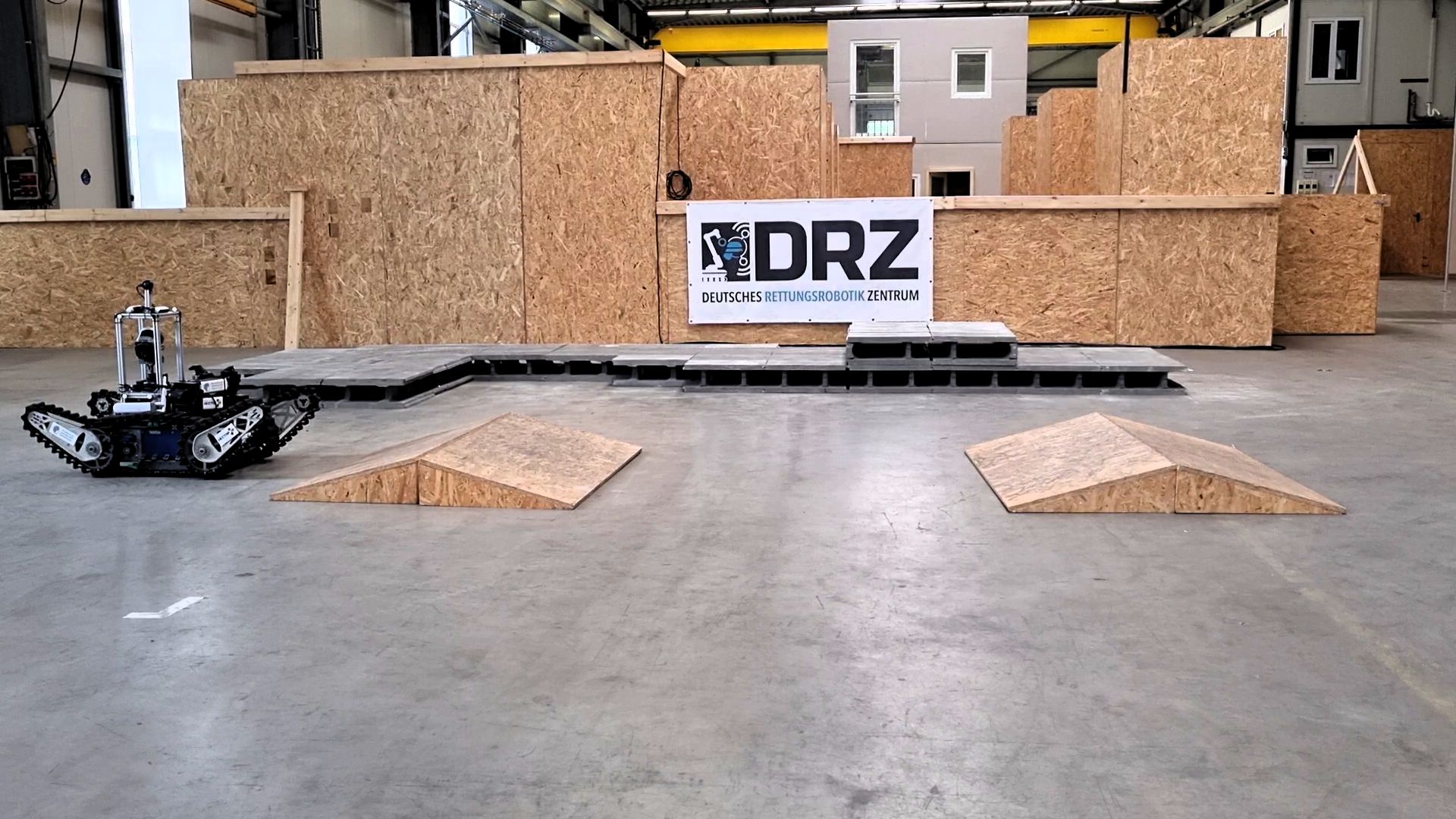}
        \caption{Continuous Ramps}\label{fig:drz_continuous_ramps}
    \end{subfigure}%
    \begin{subfigure}[t]{0.48\linewidth}
        \includegraphics[width=\linewidth]{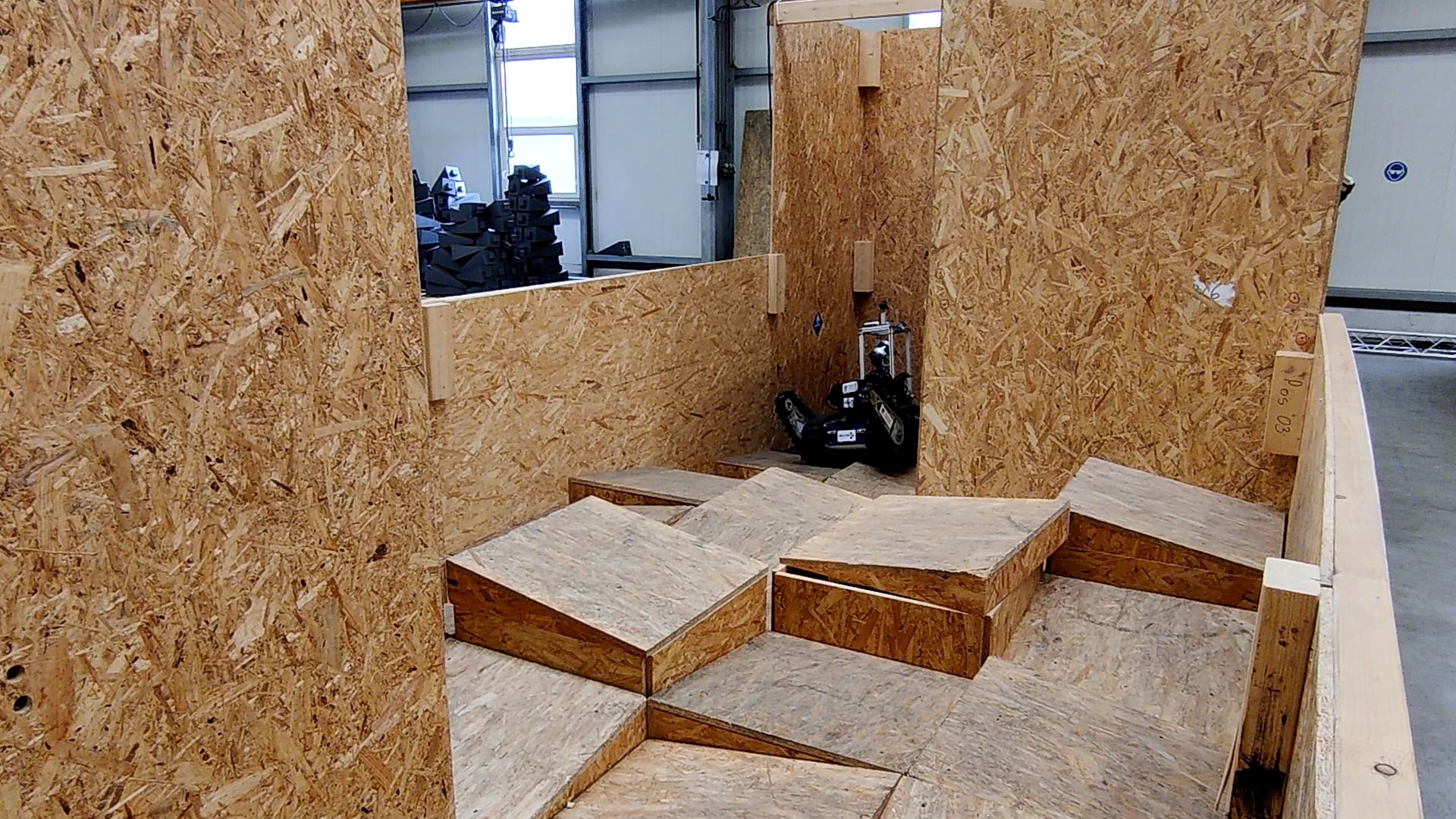}
        \caption{Elevated Ramps}\label{fig:drz_elevated_ramps}
    \end{subfigure}%
    \caption{Overview of the evaluation scenarios in the Gazebo simulator and the DRZ Living Lab. All arenas are part of the RoboCup \ac{rrl} competition.}\label{fig:evaluation_scenarios}
\end{figure}

While ground truth pose data is hard to obtain in real scenarios, it is readily available in simulation. In our evaluation, we use the Gazebo robot simulator which is based on the \ac{ode} physics engine.
Both robot platforms are driven through the four arenas (see Fig. ~\ref{fig:evaluation_scenarios}) which have been modeled in Gazebo.
The \ac{esdf} and heightmap are created from simulated depth camera data of Asterix and shared on both robots.
The results are listed in Table~\ref{tab:results}.

\begin{table}
    \centering
    \caption{Comparison between our approach and the heightmap pose prediction (HPP)~\cite{fabian2020pose} in simulation (Sim) and on the real robot (Real). The table lists the average difference between predicted and ground truth pose. Bold numbers indicate the best performance.}
    \label{tab:results}
    \begin{tabular}{l|c|c|c|c}
                       & \multicolumn{2}{c|}{\textbf{Position Error}} & \multicolumn{2}{c}{\textbf{Orientation Error}}  \\
                       & \multicolumn{2}{c|}{\textbf{Average [cm]}} & \multicolumn{2}{c}{\textbf{Average [deg]}}  \\
                       & \textbf{Ours} & \textbf{HPP} & \textbf{Ours} & \textbf{HPP} \\ \hline
        \textbf{Asterix (Sim)}        & & & &\\
        Continuous Ramps & $\mathbf{1.58}$ & $1.59$ & $\mathbf{2.89}$ & $3.29$ \\
        Curb & $1.2$ & $\mathbf{1.11}$ & $\mathbf{2.85}$ & $3.2$ \\
        Hurdles & $1.04$ & $\mathbf{0.95}$ & $\mathbf{1.05}$ & $1.42$ \\
        Elevated Ramps & $\mathbf{2.37}$ & $7.43$ & $\mathbf{6.59}$ & $12.24$ \\
        \textbf{DRZ Telemax (Sim)}        & & & &\\
        Continuous Ramps & $\mathbf{2.29}$ & $2.66$ & $\mathbf{3.38}$ & $3.98$ \\
        Curb & $\mathbf{1.93}$ & $3.02$ & $\mathbf{2.67}$ & $3.98$ \\
        Hurdles & $0.98$ & $\mathbf{0.82}$ & $1.32$ & $\mathbf{1.26}$ \\
        Elevated Ramps & $\mathbf{2.08}$ & $7.97$ & $\mathbf{4.57}$ & $10.54$ \\
        \textbf{Asterix (Real)}        & & & &\\
        Continuous Ramps & $3.06$ & $\mathbf{2.64}$ & $2.39$ & $2.39$ \\
        Elevated Ramps & $\mathbf{3.16}$ & $6.71$ & $\mathbf{5.43}$ & $11.13$ \\
    \end{tabular}
\end{table}

It can be seen that in the first three simulation scenarios, both approaches achieve similar performance on both robots, except for the "Curb" scenario on Telemax where our approach outperforms \ac{hpp} by ${\sim}\qty{1}{\centi\meter}$ in position and ${\sim}\qty{1}{\degree}$ in orientation error.
However, the strengths of our method become apparent when looking at "Elevated Ramps". This scenario is characterized by a very rough terrain with many steps and slopes. Here, our approach significantly outperforms \ac{hpp} by ${\sim}\qty{5}{\centi\meter}$ in position and ${\sim}\qty{6}{\degree}$ in orientation error on both robots.

\subsection{Motion Capture System}

While the evaluation in simulation has shown that our approach can accurately predict the robot-terrain interaction under ideal conditions, an application on the real platform is subject to higher uncertainties with additional sensor noise and model inaccuracies.
Furthermore, it is expected that the prediction quality heavily depends on the quality of the map which depends on various factors including the used sensor, surface properties, localization, and the integration algorithm.
Therefore, we also evaluate on the real Asterix robot to assess the influence of these effects and quantify the performance under realistic conditions.
Our evaluation is based on data provided by the authors of \cite{daun2021hectorgrapher}. They used the high-performance Qualisys optical motion capture system of the DRZ Living Lab\footnote{\url{https://rettungsrobotik.de/en}} to generate ground truth localization data. The \ac{esdf} and heightmap are created with a spinning VLP-16 Lidar mounted on the robot while using the tracking data of the motion capture system for scan localization.
Two scenarios are used for evaluation (see Fig. \ref{fig:drz_continuous_ramps},\ref{fig:drz_elevated_ramps}), a simplified "Continuous Ramp" scenario with two double-ramps on flat ground on a straight path and the "Elevated Ramps" which were also part of the simulated scenarios.
The results on the real platform (see bottom of Table~\ref{tab:results}) confirm the findings in simulation. While the error metrics of both approaches are very close in the easier "Continuous Ramps" scenario, our approach again significantly outperforms \ac{hpp} in "Elevated Ramps". Compared to the results in this arena in simulation, the position error increased by $\qty{0.79}{\centi\meter}$, while the orientation error even decreased by $\qty{1.16}{\degree}$. This behavior might be explained by the fact, that the simulated arena is not identical to the real one. In total, our approach achieved an average position accuracy of $\qty{3.11}{\centi\meter}$ and orientation accuracy of $\qty{3.91}{\degree}$ in this test.
These results demonstrate that our approach is robust to sensor noise and model inaccuracies and that the performance on the real robot is comparable to the simulation.

\subsection{Performance}

We measured the average time to predict a single pose on the two datasets recorded with the real Asterix platform. The computation has been performed on the mobile CPU \textit{AMD Ryzen 7 5800H} from 2021. As can be seen in the results in Table~\ref{tab:performance_results}, the average prediction time varies greatly depending on terrain complexity. However, in both cases, the timings allow for online planning applications.

\begin{table}
    \centering
    \caption{Timing results for the prediction of a single pose.}
    \label{tab:performance_results}
    \begin{tabular}{c|c|c|c}
        Scenario & Mean  & Std. Dev.  & Max. \\ \hline
        Continuous Ramps & $\qty{566}{\micro\second}$ & $\qty{214}{\micro\second}$ & $\qty{1233}{\micro\second}$ \\
        Elevated Ramps & $\qty{778}{\micro\second}$ & $\qty{282}{\micro\second}$ & $\qty{2221}{\micro\second}$ \\
    \end{tabular}
\end{table}

\subsection{RoboCup \ac{rrl} 2023 Bordeaux}
The \ac{rrl} competition is part of the annual international RoboCup event and fosters the development of mobile robots for the support of first responders in disaster situations. During the participation of Team Hector\footnote{\url{https://www.teamhector.de}} in the RoboCup World Championship 2023 in Bordeaux, the proposed approach has been applied as part of a whole-body planner~\cite{oehler2020optimization} to autonomously overcome obstacles by planning flipper trajectories online, contributing to winning the 2nd place overall as well as the "Best in Class Autonomy" and "Technology Challenge" awards.

\section{CONCLUSIONS}

In this paper, we presented a novel pose prediction approach to predict the terrain interaction of mobile ground robots with active joints. By making use of an \ac{esdf}, we benefit from its ability to represent surfaces with sub-voxel accuracy. Our evaluation has shown that this leads to a higher pose prediction accuracy in unstructured environments compared to an approach based on heightmaps.
Further research is required for environments with unstable ground where slippage or significant floor compliance might occur.

\bibliographystyle{IEEEtran}
\bibliography{IEEEabrv,literature}

\begin{thebibliography}{10}
\providecommand{\url}[1]{#1}
\csname url@rmstyle\endcsname
\providecommand{\newblock}{\relax}
\providecommand{\bibinfo}[2]{#2}
\providecommand\BIBentrySTDinterwordspacing{\spaceskip=0pt\relax}
\providecommand\BIBentryALTinterwordstretchfactor{4}
\providecommand\BIBentryALTinterwordspacing{\spaceskip=\fontdimen2\font plus
\BIBentryALTinterwordstretchfactor\fontdimen3\font minus
  \fontdimen4\font\relax}
\providecommand\BIBforeignlanguage[2]{{%
\expandafter\ifx\csname l@#1\endcsname\relax
\typeout{** WARNING: IEEEtran.bst: No hyphenation pattern has been}%
\typeout{** loaded for the language `#1'. Using the pattern for}%
\typeout{** the default language instead.}%
\else
\language=\csname l@#1\endcsname
\fi
#2}}

\bibitem{kruijff2021german}
I.~Kruijff-Korbayov{\'a}, R.~Grafe, N.~Heidemann, A.~Berrang, C.~Hussung,
  C.~Willms, P.~Fettke, M.~Beul, J.~Quenzel, D.~Schleich, \emph{et~al.},
  ``German rescue robotics center (drz): A holistic approach for robotic
  systems assisting in emergency response,'' in \emph{SSRR}.\hskip 1em plus
  0.5em minus 0.4em\relax IEEE, 2021, pp. 138--145.

\bibitem{otsu2020fast}
K.~Otsu, G.~Matheron, S.~Ghosh, O.~Toupet, and M.~Ono, ``Fast approximate
  clearance evaluation for rovers with articulated suspension systems,''
  \emph{J. Field Robot.}, vol.~37, no.~5, pp. 768--785, 2020.

\bibitem{becker20223d}
K.~Becker, M.~Oehler, and O.~Von~Stryk, ``3d coverage path planning for
  efficient construction progress monitoring,'' in \emph{SSRR}.\hskip 1em plus
  0.5em minus 0.4em\relax IEEE, 2022, pp. 174--179.

\bibitem{norouzi2017planning}
M.~Norouzi, J.~V. Miro, and G.~Dissanayake, ``Planning stable and efficient
  paths for reconfigurable robots on uneven terrain,'' \emph{JINT}, vol.~87,
  pp. 291--312, 2017.

\bibitem{oehler2020optimization}
M.~Oehler, S.~Kohlbrecher, and O.~von Stryk, ``Optimization-based planning for
  autonomous traversal of obstacles with mobile ground robots,'' \emph{Int. J.
  Mech. Control.}, no.~1, pp. 33--40, 2020.

\bibitem{elfes1989using}
A.~Elfes, ``Using occupancy grids for mobile robot perception and navigation,''
  \emph{Computer}, vol.~22, no.~6, pp. 46--57, 1989.

\bibitem{wermelinger2016navigation}
M.~Wermelinger, P.~Fankhauser, R.~Diethelm, P.~Kr{\"u}si, R.~Siegwart, and
  M.~Hutter, ``Navigation planning for legged robots in challenging terrain,''
  in \emph{IROS}.\hskip 1em plus 0.5em minus 0.4em\relax IEEE, 2016, pp.
  1184--1189.

\bibitem{oleynikova2017voxblox}
H.~Oleynikova, Z.~Taylor, M.~Fehr, R.~Siegwart, and J.~Nieto, ``Voxblox:
  Incremental 3d euclidean signed distance fields for on-board mav planning,''
  in \emph{IROS}.\hskip 1em plus 0.5em minus 0.4em\relax IEEE, 2017, pp.
  1366--1373.

\bibitem{papadakis2013terrain}
P.~Papadakis, ``Terrain traversability analysis methods for unmanned ground
  vehicles: A survey,'' \emph{Eng. Appl. Artif. Intell.}, vol.~26, no.~4, pp.
  1373--1385, 2013.

\bibitem{brunner2015design}
M.~Brunner, T.~Fiolka, D.~Schulz, and C.~M. Schlick, ``Design and comparative
  evaluation of an iterative contact point estimation method for static
  stability estimation of mobile actively reconfigurable robots,'' \emph{Robot.
  Auton. Syst.}, vol.~63, pp. 89--107, 2015.

\bibitem{jun2016pose}
J.-Y. Jun, J.-P. Saut, and F.~Benamar, ``Pose estimation-based path planning
  for a tracked mobile robot traversing uneven terrains,'' \emph{Robot. Auton.
  Syst.}, vol.~75, pp. 325--339, 2016.

\bibitem{ma2019pose}
Y.~Ma and Z.~Shiller, ``Pose estimation of vehicles over uneven terrain,''
  \emph{arXiv preprint arXiv:1903.02052}, 2019.

\bibitem{thakker2021autonomous}
R.~Thakker, N.~Alatur, D.~D. Fan, J.~Tordesillas, M.~Paton, K.~Otsu, O.~Toupet,
  and A.-a. Agha-mohammadi, ``Autonomous off-road navigation over extreme
  terrains with perceptually-challenging conditions,'' in \emph{ISER}.\hskip
  1em plus 0.5em minus 0.4em\relax Springer, 2021, pp. 161--173.

\bibitem{fabian2020pose}
S.~Fabian, S.~Kohlbrecher, and O.~Von~Stryk, ``Pose prediction for mobile
  ground robots in uneven terrain based on difference of heightmaps,'' in
  \emph{SSRR}.\hskip 1em plus 0.5em minus 0.4em\relax IEEE, 2020, pp. 49--56.

\bibitem{vsalansky2021pose}
V.~{\v{S}}alansk{\`y}, K.~Zimmermann, T.~Pet{\v{r}}{\'\i}{\v{c}}ek, and
  T.~Svoboda, ``Pose consistency kkt-loss for weakly supervised learning of
  robot-terrain interaction model,'' \emph{IEEE RA-L}, vol.~6, no.~3, pp.
  5477--5484, 2021.

\bibitem{andrew1979another}
A.~M. Andrew, ``Another efficient algorithm for convex hulls in two
  dimensions,'' \emph{Inf. Process. Lett.}, vol.~9, no.~5, pp. 216--219, 1979.

\bibitem{papadopoulos2000force}
E.~Papadopoulos and D.~A. Rey, ``The force-angle measure of tipover stability
  margin for mobile manipulators,'' \emph{Veh. Syst. Dyn.}, vol.~33, no.~1, pp.
  29--48, 2000.

\bibitem{Fankhauser2018ProbabilisticTerrainMapping}
P.~Fankhauser, M.~Bloesch, and M.~Hutter, ``Probabilistic terrain mapping for
  mobile robots with uncertain localization,'' \emph{IEEE RA-L}, vol.~3, no.~4,
  pp. 3019--3026, 2018.

\bibitem{daun2021hectorgrapher}
K.~Daun, M.~Schnaubelt, S.~Kohlbrecher, and O.~von Stryk, ``Hectorgrapher:
  Continuous-time {Lidar SLAM} with multi-resolution signed distance function
  registration for challenging terrain,'' in \emph{SSRR}, 2021, pp. 152--159.

\end{thebibliography}

\end{document}